\documentclass[letterpaper]{article} 
\usepackage{aaai20}  
\usepackage{times}  
\usepackage{helvet} 
\usepackage{courier}  
\usepackage[hyphens]{url}  
\usepackage{graphicx} 
\usepackage{graphicx}  
\usepackage{amsfonts}       
\usepackage{amsmath}
\usepackage{bm}
\usepackage{multirow}
\usepackage{xcolor}
\usepackage{mathtools}
\newcommand{\svdots}{\raisebox{3pt}{$\scalebox{.75}{\vdots}$}}

\title{Clustering and Recognition of Spatiotemporal Features through Interpretable Embedding of Sequence to Sequence Recurrent Neural Networks}
\author{Kun Su\textsuperscript{\rm 1}, Eli Shlizerman \textsuperscript{\rm 1,}\textsuperscript{\rm 2} \\ 
\textsuperscript{\rm 1}Department of Electrical \& Computer Engineering\\
\textsuperscript{\rm 2}Department of Applied Mathematics\\
University of Washington\\
Seattle, WA 98195\\
suk4,shlizee@uw.edu 
}
 
\begin{document}

\maketitle

\begin{abstract}
Encoder-decoder recurrent neural network models (RNN Seq2Seq) have achieved great success in ubiquitous areas of computation and applications. It was shown to be successful in modeling data with both temporal and spatial dependencies for translation or prediction tasks. In this study, we propose an embedding approach to visualize and interpret the representation of data by these models. Furthermore, we show that the embedding is an effective method for unsupervised learning and can be utilized to estimate the optimality of model training. In particular, we demonstrate that embedding space projections of the decoder states of RNN Seq2Seq model trained on sequences prediction are organized in clusters capturing similarities and differences in the dynamics of these sequences. Such performance corresponds to an unsupervised clustering of any spatio-temporal features and can be employed for time-dependent problems such as temporal segmentation, clustering of dynamic activity, self-supervised classification, action recognition, failure prediction, etc. We test and demonstrate the application of the embedding methodology to time-sequences of 3D human body poses. We show that the methodology provides a high-quality unsupervised categorization of movements. 
\end{abstract}

\subsection{Introduction}
Recurrent Sequence to Sequence (Seq2Seq) network models use internal states to process sequences of inputs~\cite{hochreiter1997long,luong2015effective,sutskever2014sequence,cho2014learning}. The speciality of Seq2Seq is that these models consist of encoder and decoder components. The encoder typically processes input sequences and constructs a latent representation of the sequences. In addition, the encoder passes the last internal state to the decoder as an "initialization" of the decoder. With this information the decoder transforms or generates novel sequences with a similar distribution. 
Such an architecture and its variants, e.g. attention-based Seq2Seq~\cite{cho2014learning}, showed compelling performance in applications of machine translation \cite{sutskever2014sequence}, speech recognition \cite{graves2013speech}, and human motion prediction \cite{gui2018adversarial}.

While Seq2Seq and its variants achieve strong performance on various applications, a consistent interpretation of how the encoder-decoder structure capable to embed the data for general time-series data (i.e. multi dimensional ordered sequences) and how such interpretation can be used to estimate the performance of the model is an active research topic. In this paper, we propose a dimension reduction approach to visualize and interpret representation within Seq2Seq model of general time-series data. The main contribution of this paper is to provide constructive insights on the properties which allow the encoder and the decoder components to operate optimally and to propose a low dimensional visualization method of the representation that the encoder and the decoder construct. With our embedding method we find a remarkable property of Seq2Seq: the network being trained to predict the future evolution of a sequence self-organizes the hidden units representation into separate identities (clusters or classes). The clusters are embedded in the embedding space as attractors to which embedded encoder trajectories lead. We show on human body poses data how organization of the attractors provides an unsupervised easy means for clustering sequential data into distinct identities. 

Interpretability has been an important aspect of any artificial neural network (ANN) and is expected to provide generic methodologies to assess the capabilities of different models and evaluate performance of models for given tasks.  Associating interpretability with various types of ANN is often a challenging undertaking as the typical state-of-the-art network models are high dimensional and include many components and processing stages (layers). The problem becomes more challenging when dealing with RNN sequence models and in unsupervised tasks such as synthesis and translation, in which encoder-decoder networks where found to be prevalent. In these tasks, interpretability is aimed to capture the model synthesis procedures and to demonstrate how these are being improved over training and since learning is unsupervised interpretation methodologies have the potential to provide a framework to assist with enhancing the optimization and learning.

Typically, existing interpretation methods applicable to ANNs are application oriented. For Convolutional Neural Networks (CNN), models were explained within images \cite{alain2016understanding,kim2017interpretability,zeiler2014visualizing}. For RNN, interpretation and visualization tools focus on natural language processing applications \cite{karpathy2015visualizing,collins2016capacity,foerster2017input,strobelt2019s}. Our proposed work is based and inspired by these works and extends the methodology to generic sequences, in which textual sequences are a sub-class with special time dependence (semantics).  Generic multidimensional time series are spatio-temporal sequences including nontrivial correlations in space and time. Beyond text data, there are works inspired from neuronal networks investigating the dynamics of RNN \cite{recanatesi2018signatures,farrell2019dynamic}. In our work, we aim to provide interpretation of encoder-decoder (Seq2Seq) network models for general spatiotemporal data. 

We test our methods on prediction tasks of synthetic data and of typical movements of human body joints. There are several RNN-based Seq2Seq models that achieve success on human motion prediction \cite{martinez2017human} and outperform the previous non-Seq2Seq based RNN models such as ERD \cite{fragkiadaki2015recurrent} and S-RNN \cite{jain2016structural}. Recently, Generative Adversarial Networks \cite{gui2018adversarial} have achieved better performance on this task, with the predictor network being RNN Seq2Seq.

We show that Seq2Seq optimization with gradient descent based methods forms a low dimensional embedding of internal states. The embedding can be mapped and visualized through Proper Orthogonal Decomposition (POD) of concatenated encoder and decoder internal states - the interpretable embedding. Within this embedding, the decoder evolution for each distinct sequence (decoder trajectory) is separable from other distinct sequences. Furthermore, each distinct decoder trajectory preserves both spatial and temporal properties of the sequence. The encoder trajectory initiated from various starting points connects them in the interpretable embedding space with the appropriate decoder trajectory. Monitoring the interpretable embedding space and projected trajectories in it during training shows the effect of training on data representation and assists to identify an optimal regime between under- and over- fitting. We construct synthetic data examples to demonstrate the construction of the interpretable embedding space. Next, we apply the construction of the interpretable embedding and analyze Seq2Seq performance on human joints movements datasets:H3.6M that contains $15$ different types of real body movement sequences, such as walking, eating, etc~\cite{ionescu2014human3}. To show generality and example application for the proposed approach, we apply it to CMU Motion capture dataset and perform unsupervised action recognition task.

\begin{figure}[t]
    \centering
    \includegraphics[width=0.95\linewidth]{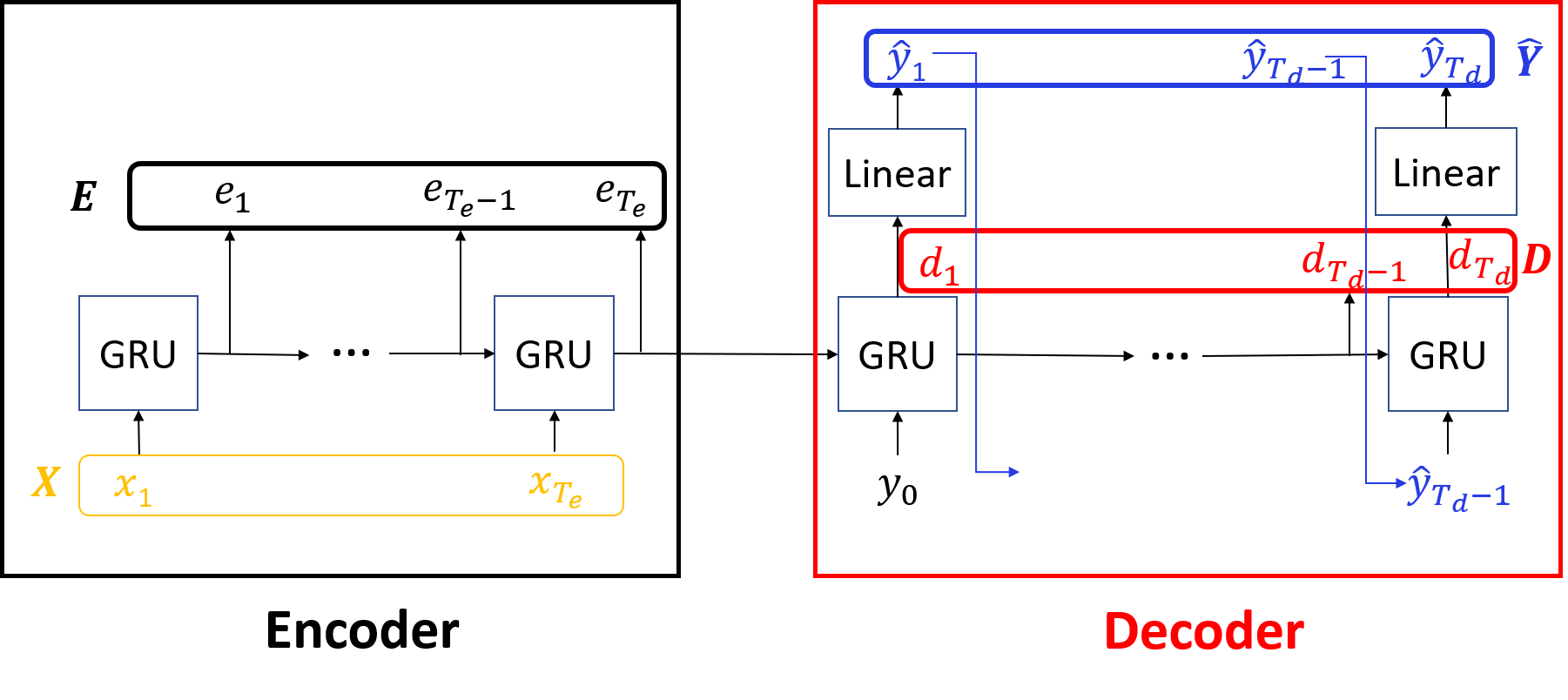}
    \caption{Seq2Seq architecture: We consider the inputs, encoder states, decoder states and outputs as spatiotemporal matrices $X,E,D,\hat{Y}$ respectively. Here the recurrent network block is Gated Recurrent Unit (GRU).}
    \label{fig:seq2seq_struc}
\end{figure}

\textbf{Setup and Spatiotemporal States of the Seq2Seq Model}:
RNN Seq2Seq model is utilized for sequence prediction (synthesis of a new sequence based on input sequence) or translation (mapping the input sequence to a new representation) (Fig.~\ref{fig:seq2seq_struc}). For general time series prediction, given a sequence of spatiotemporal data as input, Seq2Seq predicts the future sequence. We stack the sequence of input to the encoder as a matrix and call it the \textbf{encoder input matrix} ${\bf X}\in \mathbb{R}^{T_e\times M}$, where each row $x_t \in \mathbb{R}^{1\times M}$ is a time step of the input sequence at time $t$, $T_e$ is the number of time steps and $M$ is the number of dimensions in the input data. Similarly, we construct the target sequences as \textbf{target output matrix} ${\bf Y}\in \mathbb{R}^{T_d\times M}$, where $T_d$ is the number of output sequence steps to be predicted.

In addition, forward propagation of the input in RNN Seq2Seq computes the internal states of the encoder at each time step. We concatenate  and denote them as the \textbf{encoder states matrix} ${\bf E}\in \mathbb{R}^{T_e\times N}$ where each row $e_t \in \mathbb{R}^{1\times N}$ represents the states of all internal units at $t$ and $N$ is the number of hidden units. Similarly, we also define the \textbf{decoder states matrix} ${\bf D}\in \mathbb{R}^{T_d\times N}$. Typically, there is an additional fully connected linear map transforming the  decoder states to the dimensions of the output space. We denote the \textbf{decoder output matrix} ${\bf \hat{Y}}\in \mathbb{R}^{T_d\times M}$ where $\hat{y}_t$ is the predicted  $\hat{y}_t$ output at time $t$. Fig.~\ref{fig:seq2seq_struc} demonstrates the structure of  the components of RNN Seq2Seq matrices. In the figure, the encoder and the decoder are single layer GRU units that can share or not share parameters with each other depending on applications. Our approach is applicable to general types of units such as LSTMs/GRUs and number of layers. Additionally, as demonstrated, the decoder uses the output of a previous time step as an input to the current time step in both training and testing, except for the initial time step in which it receives its input from the last step of the encoder. In terms of time series prediction task, the cost function is typically defined as the MSE between \textbf{target outputs} and \textbf{decoder outputs}, $J = \frac{1}{T_d}\sum^{T_d}_{t=1}(y_t-\hat{y}_t)^2$, however, other norms or cost functions can be considered.

\textbf{Proper Orthogonal Decomposition of Spatiotemporal Matrices}: Since forward propagation within RNN Seq2Seq can be represented through spatiotemporal matrices, we propose to apply the POD method as a dimensionality reduction algorithm to construct a low dimensional interpretable embedding~\cite{shlizerman2012neural}. Specifically, we use the Singular Value Decomposition to decompose the matrices into orthogonal spatial modes (PCs) and time dependent coefficients and singular values (scaling) associated with each mode. Particularly, given a matrix $A\in\mathbb{R}^{T\times N}$ we center the matrix around the origin, obtaining $A_c$, and then apply SVD such that $A_c = U\Sigma V^T,$ where $U\in\mathbb{R}^{T\times T}$ is an orthogonal matrix of time dependent coefficients, $\Sigma\in\mathbb{R}^{T\times N}$ is the matrix of singular values and $V\in\mathbb{R}^{N\times N}$ is the matrix of spatially dependent components. To determine the number of PCs, we compute the singular value energy (SVE), $\text{SVE} = \sum^n_{i=1}\sigma^2_i$ where $\sigma_i$ is singular value corresponding to PC mode $i$. We compute the number of modes such that $90\%$ or $99\%$ of the energy is retained ($\text{SVE}_k = \sum^k_{i=1}\sigma^2_i, \frac{\text{SVE}_k}{\text{SVE}}\leq p$ where $k$ is the number of modes and $p$ is the percentage). With the number of dominant spatial features, we can truncate $A_c$ by projecting it onto the PC modes to get a low dimensional matrix  $A_{\text{(PC=n)}}\in\mathbb{R}^{T\times n}$, where (PC=n) denotes $n$ principal components are retained. If we choose $n=2 \text{ or } 3$, we can visualize the representation in 2D or 3D. The axes are the orthogonal PC modes.

\textbf{Clustering}: While visualization of projected dynamics could be informative~\cite{maaten2008visualizing}, 2D or 3D visualized dynamics do not reveal the intricacies in the representation of different datasets. In particular, here we would like to evaluate the separability of projections of distinct trajectories in the interpretable embedding space.  We thereby propose to augment the embedding with clustering approaches such as \textbf{K-means++}, an extended version of the standard K-means algorithm~\cite{arthur2007k} or \textbf{agglomerative} clustering, a bottom up approach of hierarchical clustering. Since the number of trajectories and time steps are known, we can use the \textbf{Adjusted Rand Index (ARI)} to evaluate the clustering performance.

\subsection{Interpretable Embedding for Seq2Seq Networks}
\begin{figure}[t]
    \centering
    \includegraphics[width=\linewidth]{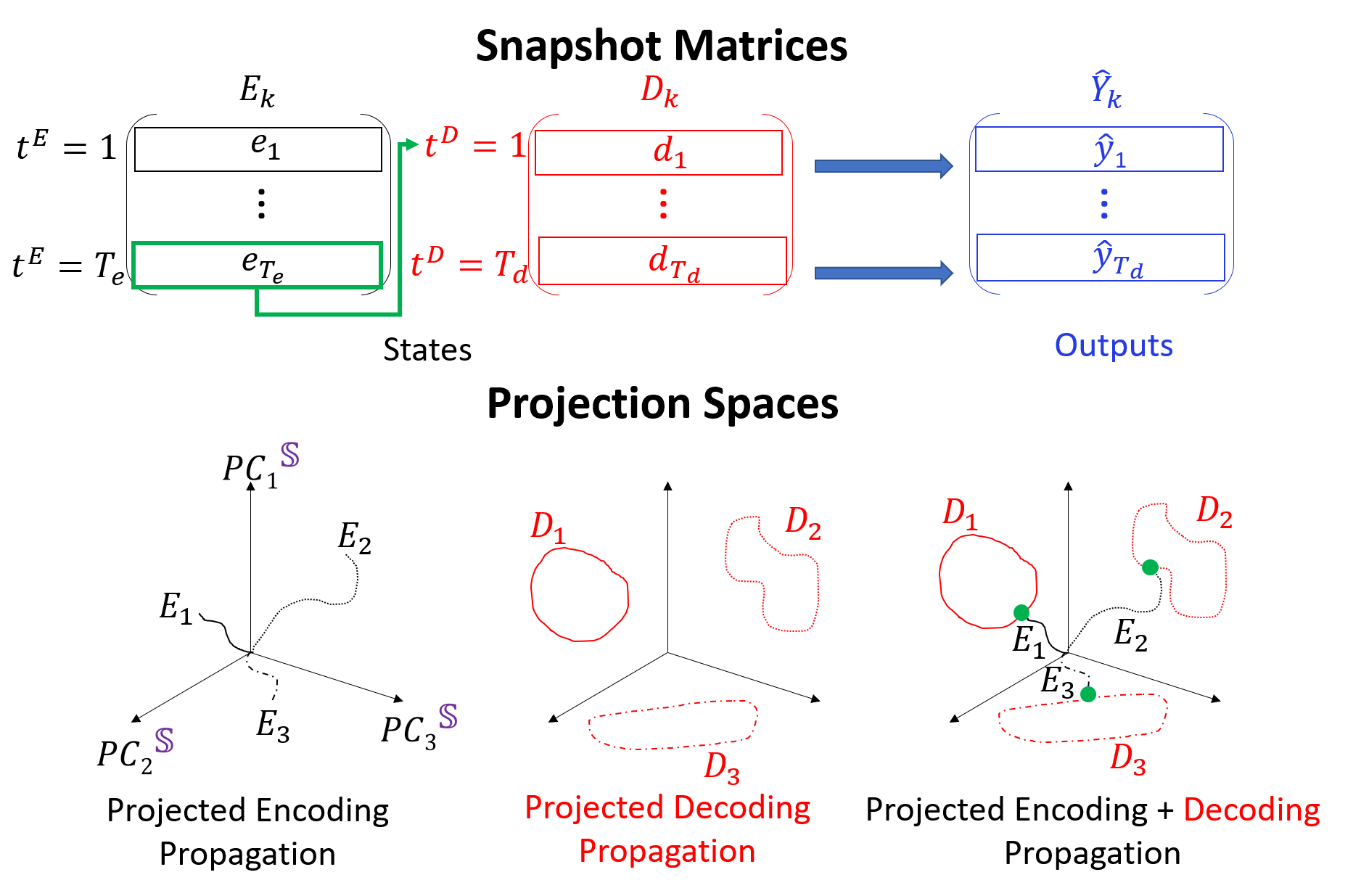}
    \caption{POD is performed on $E_k$ and $D_k$, obtained for each type of data $k$ and stacked to $\mathbb{S}$. Three distinct trajectories are shown in the low dimensional embedding space. The encoder trajectories (black) start from the origin and diverge in different directions. The decoder trajectories (attractors; red) are hence placed in separate locations in the space. The last point of each encoder trajectory (green) connects the encoder and the decoder trajectories.}
    \label{fig:seq2seq_interpret}
\end{figure}

The generic goal of RNN Seq2Seq model is to continue (predict) the evolution of each given input sequence chosen from a (test) dataset which includes $K$ different types of multidimensional time series. Such a goal is challenging as it requires the network to generate a sequence which superimposes the typical dynamics of that particular type and the individual dynamics of the given tested input sequence. We show that the methods described above can be applied to construct an interpretable embedding for the RNN Seq2Seq model depicted in Fig.\ref{fig:seq2seq_interpret}.

To construct the embedding space basis we concatenate the matrices ${\bf E}$ and ${\bf D}$ for each single forward propagation into a \textbf{state matrix} 
\begin{align*}
        {\bf S} = \begin{bsmallmatrix}
    E\\
    D
    \end{bsmallmatrix} \in \mathbb{R}^{(T_e+T_d) \times N}, ~~~
        {\bf \mathbb{S}}=\begin{bsmallmatrix}
    S_{1}\\
    \svdots\\
    S_{K}
    \end{bsmallmatrix}.
\end{align*}  Concatenation of forward propagation evolution for all considered time series in the dataset will result with a \textbf{global state matrix} denoted as
${\bf \mathbb{S}}$.
POD application on ${\bf \mathbb{S}}$ provides the PC modes which are the axes of the interpretable embedding space. Dimension reduction of the embedding space is performed by considering only $n$ PCn modes which singular values are included in representation of particular total SVE (e.g. $90\%$ or $99\%$) and truncating the rest of the modes.

To inspect the propagation of single time series through the network, we can project the matrices ${\bf E}$, ${\bf D}$ or ${\bf S}$ onto the low dimensional space spanned by PCn modes. As we show below, we find that the dimension of the embedding space can be very low even for high dimensional data. We depict the structure of such projections onto PC3 embedding space in Fig.~\ref{fig:seq2seq_interpret} bottom.  Encoder trajectories (black, left) start from initial points projected to the embedding space (we choose initial states as zeros and therefore the trajectories are always initiated at the origin) and evolve in different directions in the space. Decoder trajectories (red, middle) appear as attractors in the space and clustering approaches are used to determine (e.g. K-means) how separable they are in the space. The encoder and the decoder connect via a single time step which corresponds to a point (green, right). Composition of the three types of projections corresponds to an interpretation of the propagation in RNN Seq2Seq. In particular, we show that the encoder trajectory takes the sequence from an initial point and evolves it to the corresponding starting point on the decoder trajectory (we call it attractor or cluster). The decoder continues the evolution from there. As we show below, it appears that the gradient descent training succeeds to organize the decoder attractors in the embedding space such that they are easily clustered. Such an arrangement explains the uniqueness of RNN Se2Seq training in which the cost function minimizes the error between the decoder output and the actual output and there is no minimization on the encoder. Therefore, the decoder is trained to predict different features for different types of inputs (training to optimize clustering of types of data and capture unique features) with the encoder trajectory (and not only the last time step of it) being a sequential constraint that connects the cluster to the initial state. In practice, the longer the encoding sequence the better is the prediction (clustering property) and such an interpretation is evident from the low dimensional embedding space as well.

It is possible to monitor the the construction and the changes in the embedding space and projected trajectories within it during training. Specifically, at each training iteration $i$, for each type of data, we construct the matrices $E^{i},D^{i},\mathbb{S}^i$ and perform the aforementioned procedures.

\subsection{Interpretable Embedding Space Applied to Synthetic Data}
We first create two types of synthetic 2D trajectories which follow a circumference of (i) a unit circle $X_{circle}$ (ii) an ellipse $X_{ellipse}$. The encoder input is binned into $T_e$ time steps such that $X_{circle}, X_{ellipse} \in \mathbb{R}^{T_e\times 2}$ with  $T_e = 50$ such that it completes a full period. Given full circle or ellipse dynamics as the encoder input, the objective is to predict another full circle or ellipse by the decoder, i.e., the target sequence $Y$ is expected to be the same as the input sequence ($Y\equiv X$) and the \textbf{decoder output matrix} $\Hat{Y}$ has the same dimension as $Y$ and $X$. The cost function $J = \frac{1}{T_d}\sum^{T_d}_{t=1}(y_t-\hat{y_t})^2$ is the MSE between the target and the prediction with $T_d=T_e$. For both the encoder and the decoder we use a single layer GRU with $16$ neurons ($E$,$D\in \mathbb{R}^{T_e\times 16}$) and the ADAM optimizer to train the model for $5000$ iterations where the cost function almost converges to zero such that the model can predict the trajectory with high accuracy.

\begin{figure}[t]
    \centering
    \includegraphics[width=0.9\linewidth]{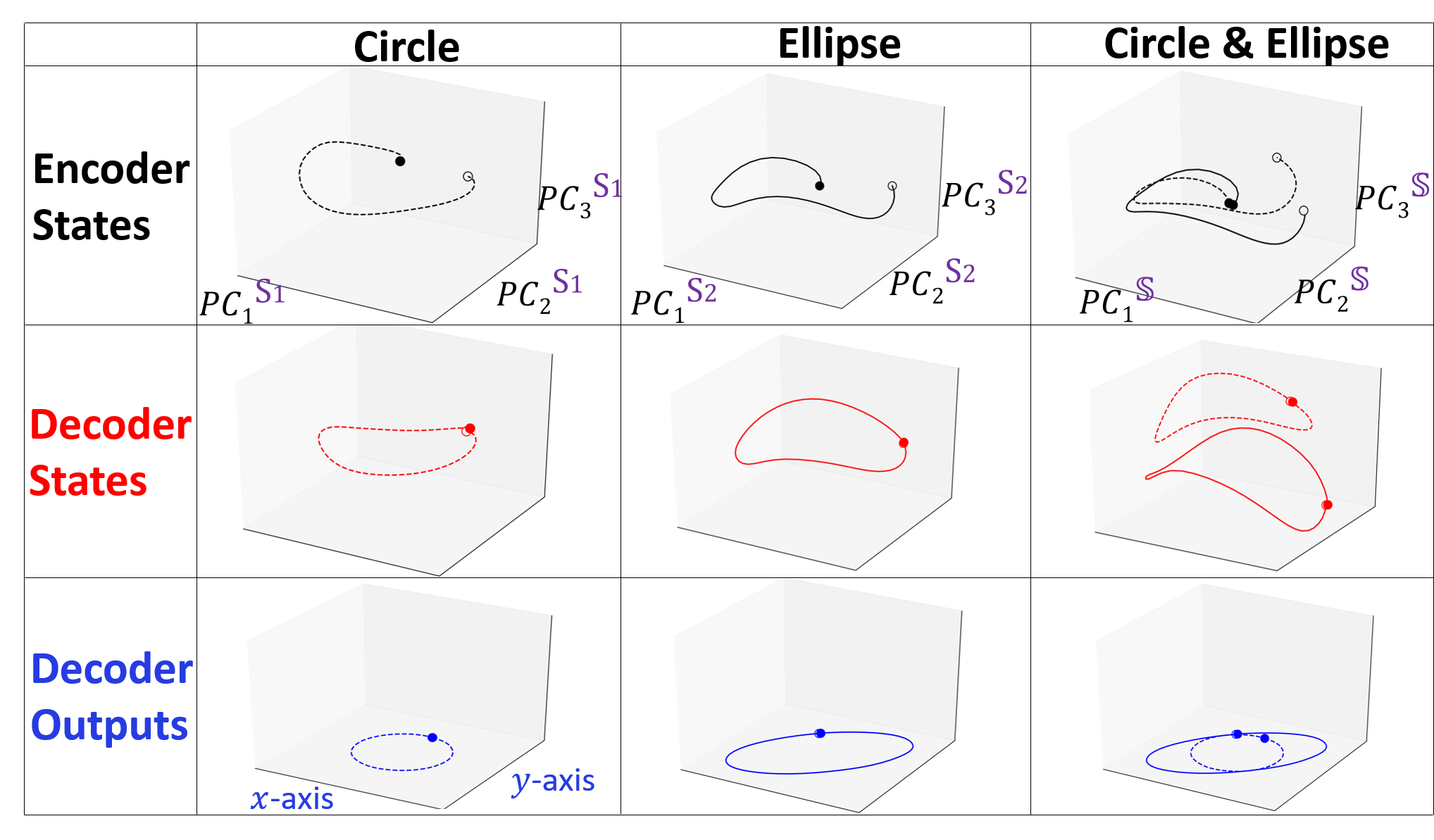}
    \caption{Projected trajectories in the embedding space spanned by PC3 for the encoder states, decoder states and decoder outputs matrices. Dashed and solid lines represent $X_{circle}$ and $X_{ellipse}$ respectively. Opaque and transparent points denote the starting and the ending points of the trajectory.}
    \label{fig:synthetic_flow}
\end{figure}
We show in Fig.~\ref{fig:synthetic_flow} the projections of forward propagation in RNN Seq2Seq trained on (i) circle (ii) ellipse (iii) both circle and ellipse. In each model, we obtain the PC3 embedding space from the matrix $\mathbb{S}$ and examine the projections of the encoder states matrix ${\bf E}$, decoder states matrix ${\bf D}$ onto the space and also show the decoder output matrix ${\bf \hat{Y}}$ projected onto $x-y$ space. We observe that the projections of the encoder and the decoder states are deformed and not necessarily preserve the same form as the input and the output, however, linear transformation of the decoder deforms the trajectory to be correctly represented in the $x-y$ space. Notably, all encoder projections start near the origin since our initial states are zero by default.

The last state of the encoder in NLP applications is typically considered to contain the information of all previous states, an assumption that gives rise to the attention-based Seq2Seq model. However, our results indicate that the full encoder sequence is important and the last encoder state is simply providing a starting point for the decoder to continue. Such an effect of the encoder is easily observed in considering continuous time series (as we show in prediction of human movement data) and deviates from the interpretation of the encoder role in textual semantic sequences. 

We focus on the model trained on both the circle and the ellipse to better understand how RNN Seq2Seq is able to predict both spatio-temporal series in an unsupervised way without any guidance. Projections onto the PC embedding space (right column of Fig.~\ref{fig:synthetic_flow}). Encoder projections indicate that the two trajectories corresponding to distinct types of output sequences, start from points near the origin however diverge as the sequence evolves and end up at more distant points. The projected trajectories of the decoder states start from these distinct points and continue to perform prediction in completely separable shapes. Effectively we observe that the decoder states projections are clustered in the embedding space. Application of agglomerative clustering and cosine similarity indeed verify these observations.

\begin{figure}[t]
    \centering
    \includegraphics[width=0.9\linewidth]{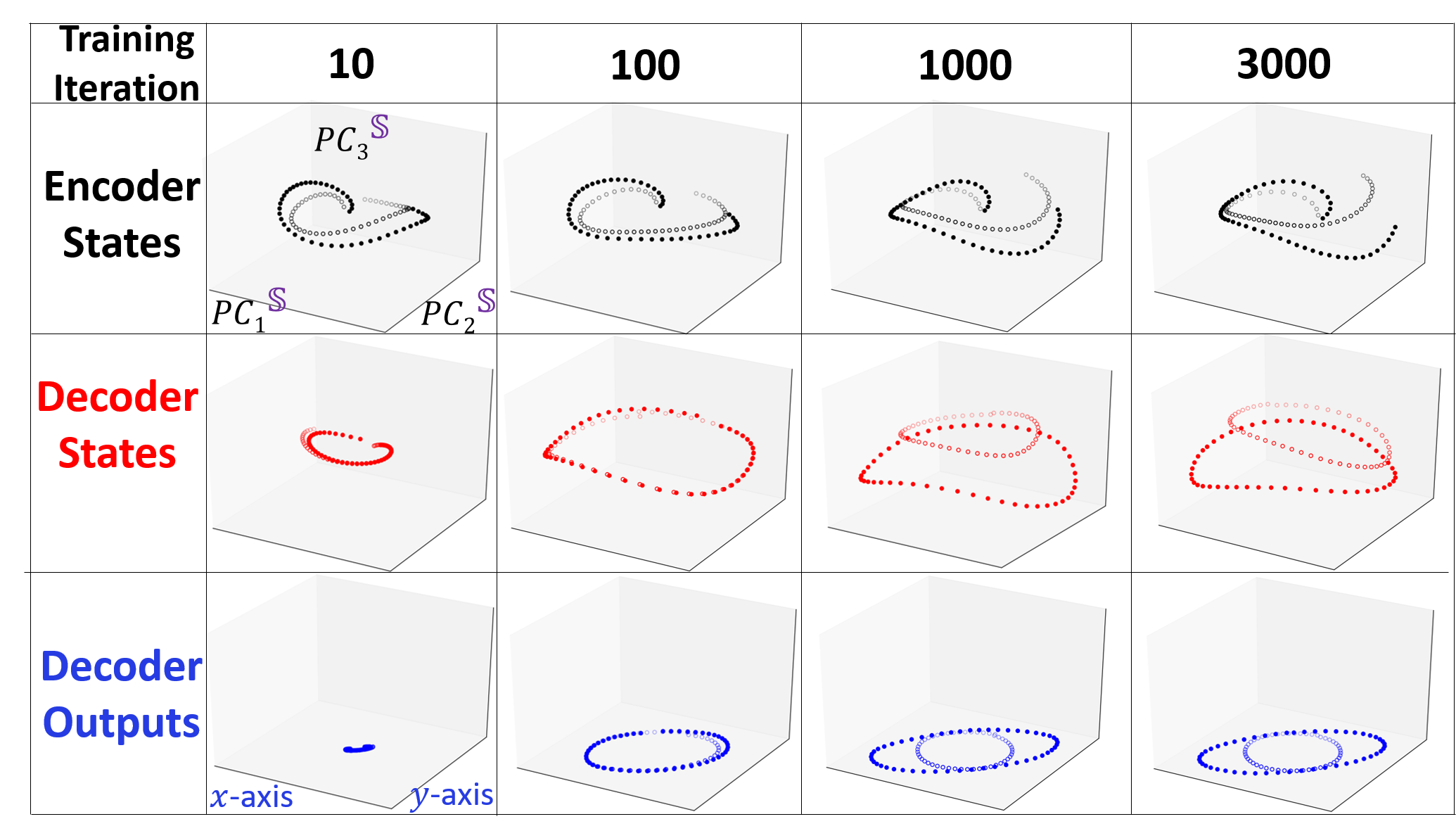}
    \caption{Snapshots of the evolution of trajectories projected to the interpretable embedding space when they undergo training. We use similar line identifiers and colors as in Fig. \ref{fig:synthetic_flow}. At the beginning of the training, two trajectories are very similar and positioned close to each other. As training proceeds the two type of trajectories emerge and separate from each other.}
    \label{fig:syn_evo_cluster}
\end{figure}

To visualize how Seq2Seq learns to differentiate the two shapes with training, we keep track of the evolution of states representation and depict in Fig.~\ref{fig:syn_evo_cluster} the three projections as in Fig.~\ref{fig:synthetic_flow} at iterations $i=10,100,1000,3000$.
In the beginning of training ($i=10$), all projections of the two shapes are similar. Projections of the decoder states appear to be distinct for a few several initial points but when the sequence evolves forward, the trajectories appear to converge to the same point. When the model undergoes additional training, after $i=100$, Seq2Seq appears to have learned a single pattern (ellipse like), however, is unable to generate two distinct predicted shapes. At  $i=1000$ the two trajectories separate and obtain accurate predictions at the same time. The evolution during training reveals that the model learns one general pattern first and then gradually evolves into two different separable patterns. 

\begin{figure}[t]
    \centering
    \includegraphics[width=0.8\linewidth]{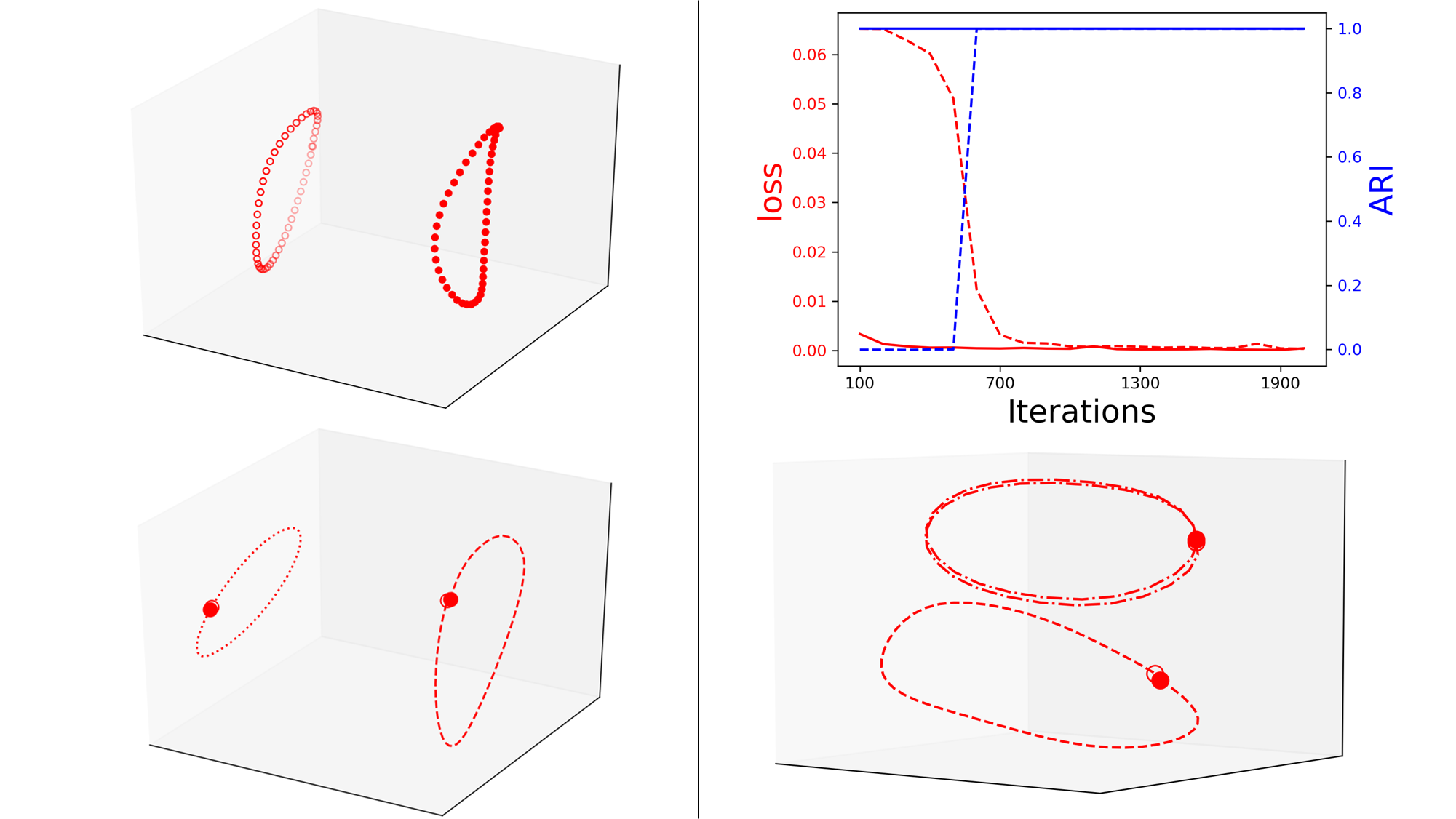}
    \caption{Top: Left: Decoder states embedded projection with one-hot encoding; Right: The inverse correlation between loss (loss) and clustering performance (blue) of unit circle and ellipse with one-hot encoding (solid line) and without (dashed line). Bottom: Left: decoder representation of centered (dashed) and shifted (dotted) circles. Right: decoder representation of same circle with different rotation frequency: $Te=50$ (dashed) and $Te=25$ (dashed-dotted).}
    \label{fig:loss_vs_cluster}
\end{figure}

Notably, during training the loss is inverse proportional to the clustering performance (ARI). This means that with training the RNN Seq2Seq model learns to recognize the type of input sequence. We conjecture that separability is correlated with successful training and test the conjecture by repeating the same prediction task with additional one-hot encoding label for the circle and the ellipse. Indeed, with the additional labels, the model learns these two sequences much faster and they appear separable in the embedding space (Fig.~\ref{fig:loss_vs_cluster} top).
We also verify that RNN Seq2seq model can encoder both spatial and temporal features in the embedding space. For spatial features, we train the model to learn centered and shifted circles. While both trajectories are circles, RNN Seq2seq and the projections of its decoder states to the embedding space distinguish the trajectories well (Fig.~\ref{fig:loss_vs_cluster} bottom left). For temporal features, we train the network to predict two centered unit circles, sampled with different rates of $T_e = 50$ and $T_e = 25$ (different rotation speeds on the circle). While both circle trajectories coincide in $x-y$ space, RNN Seq2seq and the projections of its decoder states to the embedding space result with separable attractors with apparent temporal feature difference between the first and second cycles (Fig.~\ref{fig:loss_vs_cluster} bottom right). These investigations help us to conclude that the embedding space can be effectively used with clustering to represent and assist in evaluation of different types of learned spatial and temporal features.

\subsection{Human Body Joints Movements Data}
To test the proposed interpretable embedding methodology on realistic data, we use the Human 3.6 million (H3.6M), which is currently one of the largest publicly available data sets of motion capture data~\cite{ionescu2014human3}. It includes $7$ actors performing $15$ various activities such as walking, sitting, and posing. Each movement is repeated in $2$ different trials. We use different people for training and testing; $6$ of the actors motion as the training set and the other actors motion as a testing set. Human pose is represented as an exponential map representation of each joint, with a special pre-processing of global translation and rotation. The number of features of body joints dynamics is $M=54$. For consistency with the previous synthetic example, we provide $T_e=50$ frames of input data and predict next $T_d=50$ frames. As a result, the \textbf{actual input matrix}, \textbf{actual output matrix} and \textbf{decoder output matrix} will be $X,Y,\Hat{Y}\in \mathbb{R}^{T_d\times54}$. 
Seq2Seq model was shown to be successful in such a prediction task~\cite{martinez2017human}. We use a similar setup with a single layer GRU of $N=1024$ units for both the encoder and the decoder. The \textbf{encoder and decoder states matrices} and are $E,D\in \mathbb{R}^{T_e\times N}$. The cost function is the mean square error between the ground truth and the predicted output $J = \frac{1}{T_d}\sum^{T_d}_{t=1}(y_t-\hat{y_t})^2$. We use a batch size of $B = 15$ such that every training iteration can contain all actions. We train the model with various gradient descent based optimizers and show our results for ADAM (the fastest converging optimizer for this model).
Here we show results with a similar network setup as in previous work, however, our investigations indicate that our method is applicable to other variants of RNN Seq2seq model (e.g. non-parameter sharing RNNs, multi-layer RNNs, etc).
\begin{figure}[t]
    \centering
    \includegraphics[width=0.9\linewidth]{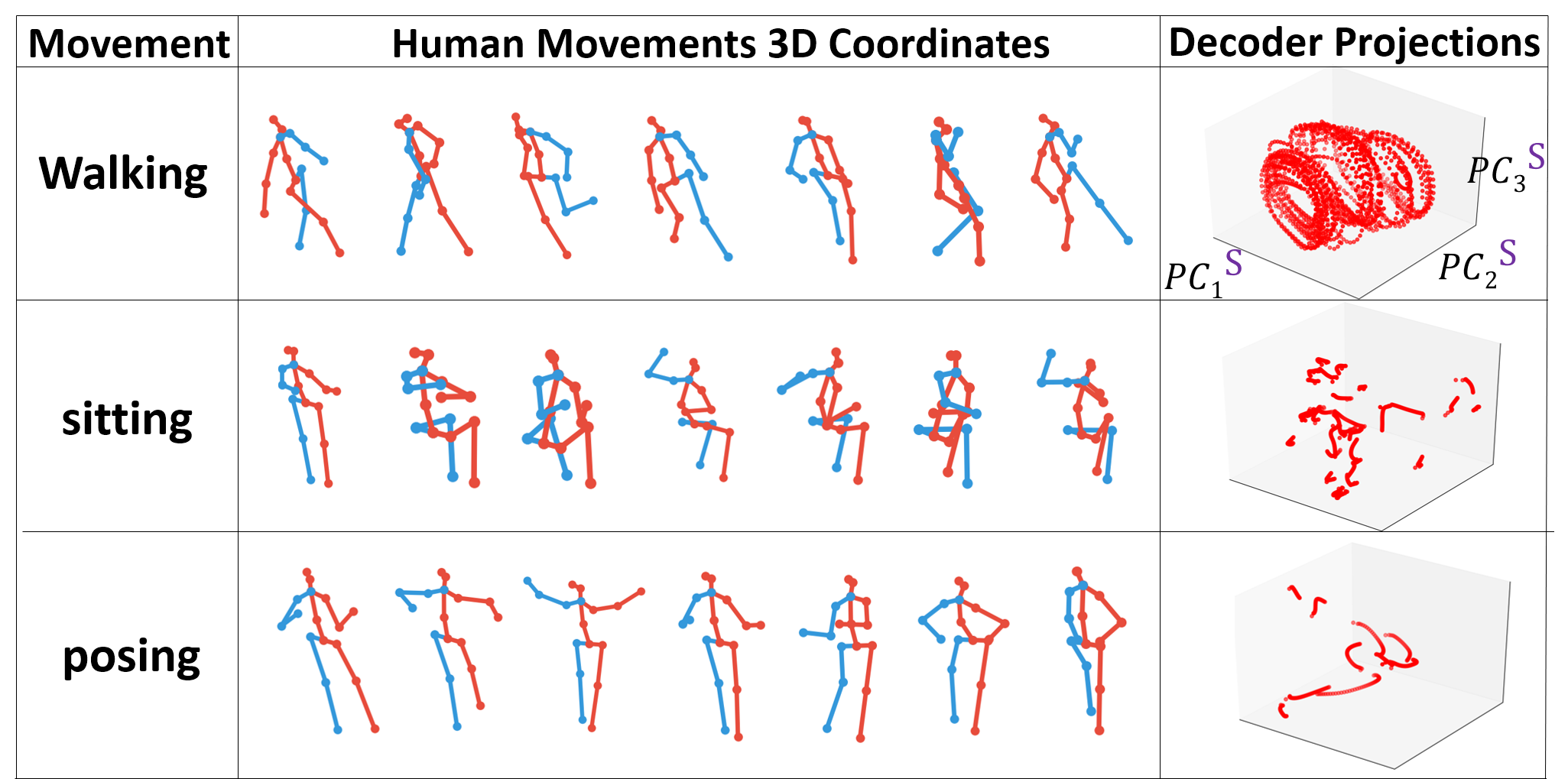}
    \caption{Different human joint movement types and their corresponding decoder states projected to the interpretable embedding space.}
    \label{fig:long_predict}
\end{figure}

\begin{figure}[t]
    \includegraphics[width=0.9\linewidth]{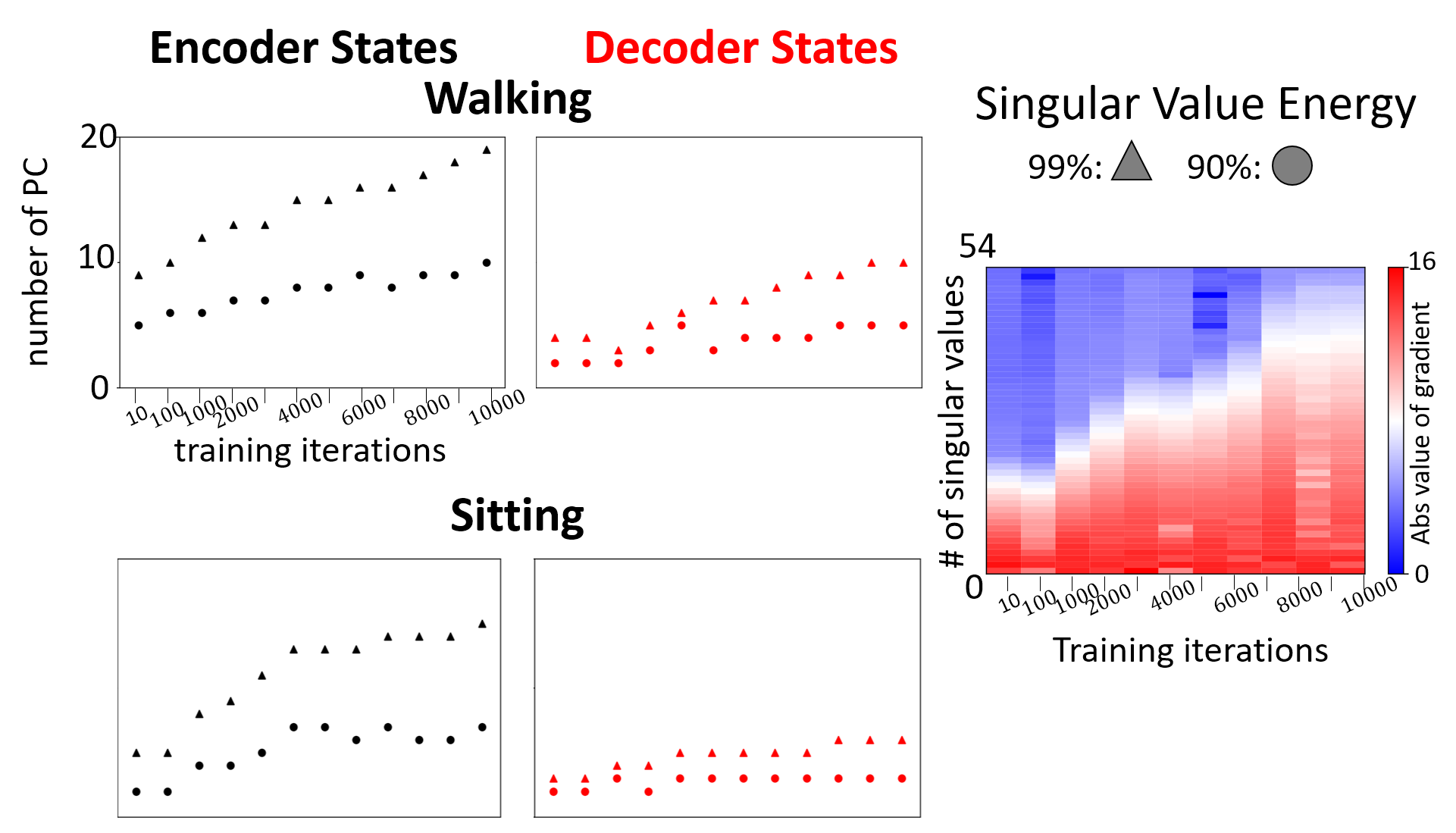}
    \caption{Left: Examples of evolution of the number of dominant modes to reach $90\%$ and $99\%$ in encoder and decoder for walking and sitting,respectively. Right: evolution of absolute changes of singular values along the training for "walking" decoder states matrix.}
    \label{fig:evo_singular}
\end{figure}
We visualize the projections onto PC3 embedding space for every type of action. In order to see the pattern for the full action, we train the model for prediction and then continuously perform forward propagation on encoder and decoder sequences in a sliding a window starting from the very beginning.
In Fig.~\ref{fig:long_predict}, we show examples of the joints evolution (connected with lines) alongside with the decoder states projection onto PC3 embedding space. We clearly observe that each individual action has its own trajectory evolution (attractor) pattern even in 3D. For example, as expected, walking data corresponds to periodic circular pattern with rather fixed period and actions such as sitting correspond to non-periodic trajectories in the embedding space. To get a better understanding on what should be the appropriate dimensions of the embedding space we monitor the number of dominant PC modes to reach $90\%$ and $99\%$ SVE for encoder and decoder matrices POD for every type of action during training (Fig.~\ref{fig:evo_singular}).

We observe a low number of dominant modes ($< 10$ and $<20$ on average for $90\%$ and $99\%$ respectively) needed to span most of the energy. The number of modes increases with training and the requirement of the energy threshold (accuracy of the embedding). The rate of the increase depends on the type of movement. For example, as shown in Fig.~\ref{fig:evo_singular}, the number of  dominant modes for "sitting" is much smaller than for walking. Furthermore, the number of dominant modes required to reach $90\%$ SVE increases more slowly than numbers of modes to reach $99\%$. These results indicate that the model learns additional details with training. However, most of the main features captured in $(90\%)$ are learned in the first iterations. Such an observation is supported by analyzing the gradients of the singular values during training (Fig.~\ref{fig:evo_singular}, right). As training proceeds, additional singular values are gradually being optimized. Notably, we observe that encoder states would have more modes than decoder states. The reason stems from the encoder trajectory starting from the origin and connecting to the decoder attractors in various parts of the embedding space. Such trajectories are hence including  mixed characteristics of the attractor and of the path to it resulting in more irregular trajectories requiring additional modes to be represented.
\begin{table}[h]
\scriptsize
\centering
\caption{Clustering results}
\label{tab:cluster}
\begin{tabular}{|c|c|c|c|}
\hline
                                  & Training iterations    & Dimension & ARI (\%)             \\ \hline
\multirow{6}{*}{Encoder states}    & \multirow{2}{*}{10}    & dim=3     & 52            \\ \cline{3-4} 
                                  &                        & dim=1024  & 61              \\ \cline{2-4}
                                  \cline{2-4} 
                                  & \multirow{2}{*}{4000}  & dim=3     & 56            \\ \cline{3-4} 
                                  &                        & dim=1024  & 79              \\ \cline{2-4}
                                  \cline{2-4}
                                  & \multirow{2}{*}{10000} & dim=3     & 59              \\ \cline{3-4} 
                                  &                        & dim=1024  & 86              \\ \hline\hline
\multirow{6}{*}{{\color{red} Decoder states}}    & \multirow{2}{*}{{\color{red} 10}}    & {\color{red} dim=3}     & {\color{red} 29}             \\ \cline{3-4} 
                                  &                        & {\color{red}dim=1024}  & {\color{red} 32}             \\ \cline{2-4}       \cline{2-4}
                                  & \multirow{2}{*}{{\color{red} 4000}}  & {\color{red} dim=3}     & {\color{red} 97}            \\ \cline{3-4} 
                                  &                        & {\color{red} dim=1024}  & {\color{red} \textbf{100}} \\ \cline{2-4} 
                        \cline{2-4} 
                                  & \multirow{2}{*}{{\color{red} 10000}} & {\color{red} dim=3}     & {\color{red} 94}            \\ \cline{3-4} 
                                  &                        & {\color{red} dim=1024}  & {\color{red} \textbf{96}}  \\ \hline \hline
\multirow{2}{*}{{\color{blue} Joints data}} & \multirow{2}{*}{}      & {\color{blue} dim=3}     & {\color{blue} 59}          \\ \cline{3-4} 
                                  &                        & {\color{blue} dim=54}    & {\color{blue} 80}         \\ \hline
\end{tabular}
\end{table}

To understand how Seq2Seq forms distinct attractors and differentiates various actions, we construct matrices ${\bf E},{\bf D},{\bf \hat{Y}}$ for all $15$ actions for which the model was trained. We apply the K-means++ clustering to $\mathbb{D}$ to evaluate the separability of Seq2Seq in the interpretable embedding space and the dimension of the space which provides efficient clustering property. (Table~\ref{tab:cluster}). 

\begin{figure}[t]
    \centering
    \includegraphics[width=\linewidth]{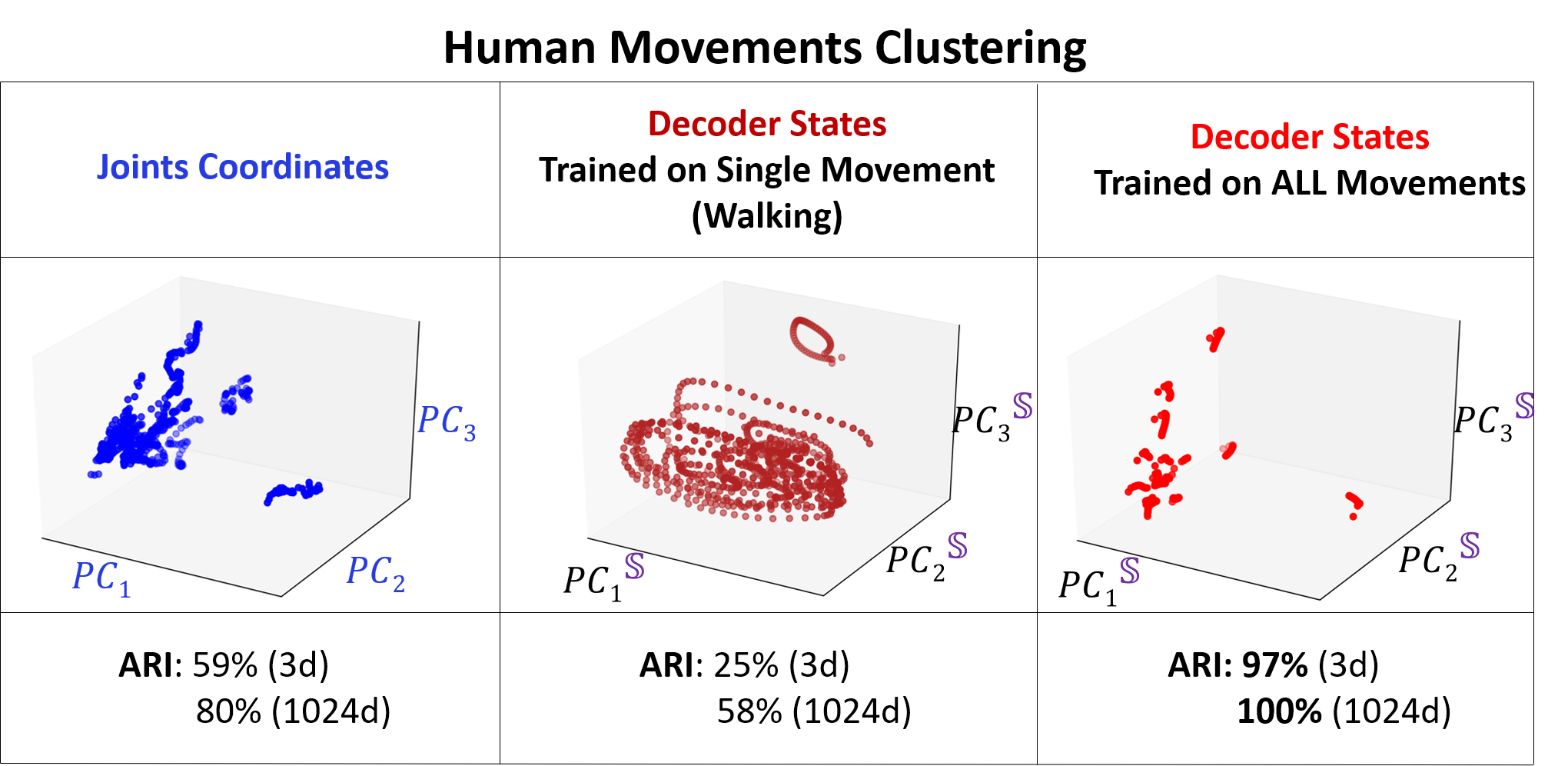}
    \caption{Visualization of clustering in 3D. Left: Joints coordinates projected data, Middle: Overfitting for walking data. Right: Decoder states at training iterations=2000.}
    \label{fig:cluster}
\end{figure}

We compare clustering of the encoder and the decoder trajectories with respect to ARI at different training iterations and for different dimensions (dim$=3$ and dim$=1024$) of the embedding space. In addition, we cluster the body joints data (in dim$=3$ and dim$=54$). Only clustering of the decoder attractors is able to reach $100\%$ clustering in both measures for dim$=1024$ of the embedding space. The next best clustering performance is for the decoder attractors in dim$=3$ ($97\%$ ARI which is significantly higher than joints data for dim=$3$ and dim=$54$) at iterations $=4000$. After this number of iterations the attractors start to approach each other instead of diverging. Encoder trajectories are not clustered well in any dimension and their clustering does not significantly change with training. In Fig.~\ref{fig:cluster} we visualize the clusters (for joints coordinates, overfitted trained model, trained model on all movements) in 3D. We show that even visually, the decoder attractors are scattered in the space indicating  almost perfect clustering property of PC3 space. If we keep adding the number of PCs to $10$, it would be enough to achieve the perfect clustering result.

\begin{figure}[t]
    \includegraphics[width=\linewidth]{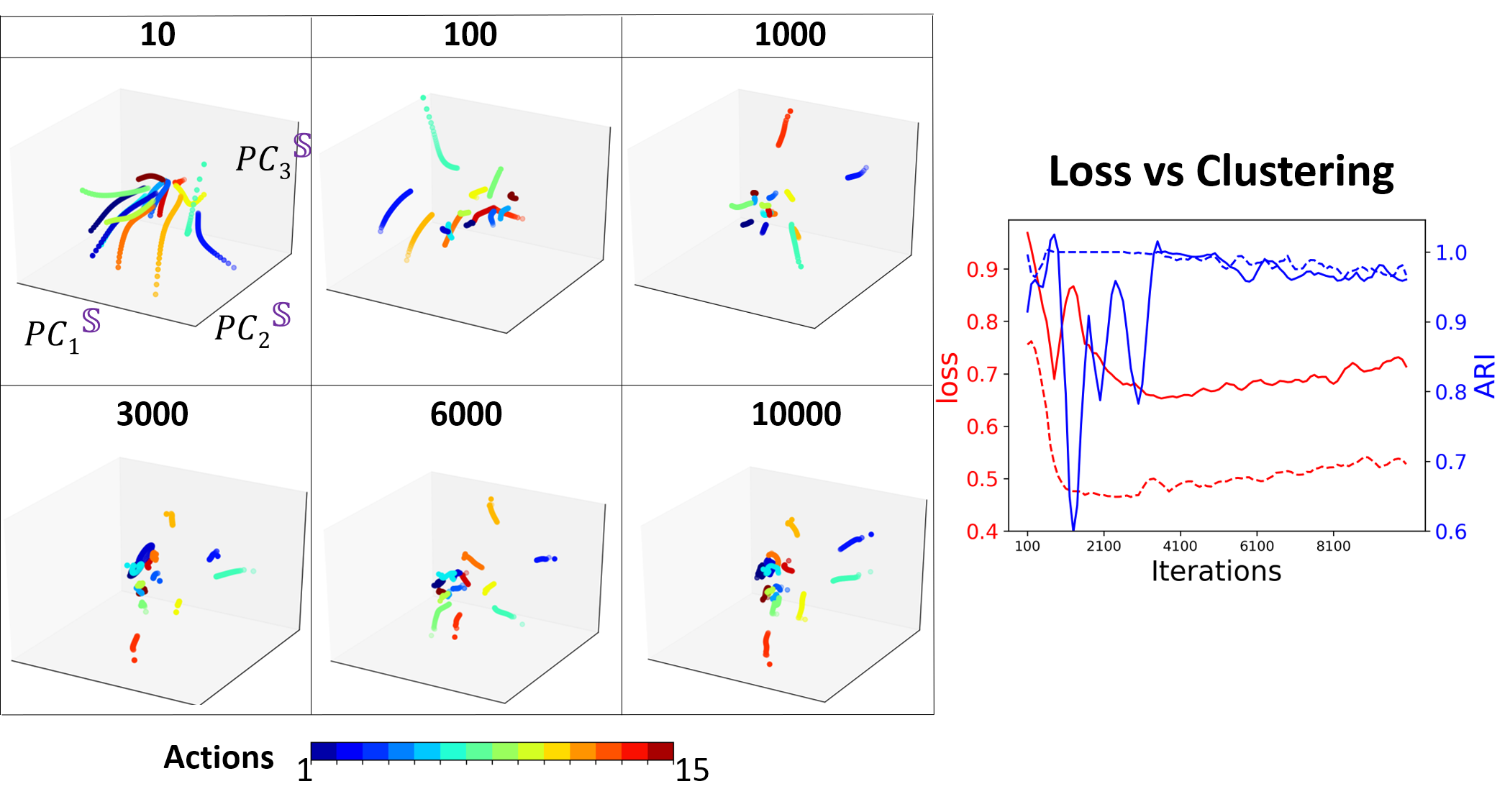}
    \caption{Left: projection of $\mathbb{D}$ on the basis of $\mathbb{S}$ at different training iterations, each color represents an action type. Right: comparison of loss (red) with clustering (blue) during training, with one-hot encoding (dashed) and without one-hot encoding (solid).}
    \label{fig:evo_pc}
\end{figure}

To understand how training shapes the decoder attractors and their clustering, we mark each distinct movement attractor by a different color and monitor their representation in PC3 embedding space for various training iterations (Fig.~\ref{fig:evo_pc}). One of our key observations is that the \textbf{clustering performance} reaches a \textbf{peak} after considerable training (~$4000$ iterations) which is at the same time that the \textbf{validation loss} reaches its \textbf{minimum} point. After that, the validation loss increases as clustering becomes inferior. Such performance is known as over-fitting eventually will lead to a model performing on certain actions only. We demonstrate such a case by training Seq2Seq with more "walking" action data and then test the model with all actions. Indeed, no matter what type of action is given as an input, the decoder states will always be similar to a circular trajectory that represents the "walking" action. Hence, we propose that the \textbf{clustering property} of the embedding space could be used as an indication for \textbf{sufficient and optimal training} of the model. To show the importance of the separability of clusters, similar to synthetic case, we compare the results with one-hot encoding data. We find that the model converges much faster and achieves stable high clustering performance very early. In both cases, clustering performance is strongly correlated with clustering, i.e., when overfitting starts to occur, clustering performance starts to deteriorate.

\subsection{Unsupervised action recognition}
\begin{figure}[t]
    \centering
    \includegraphics[width=\linewidth]{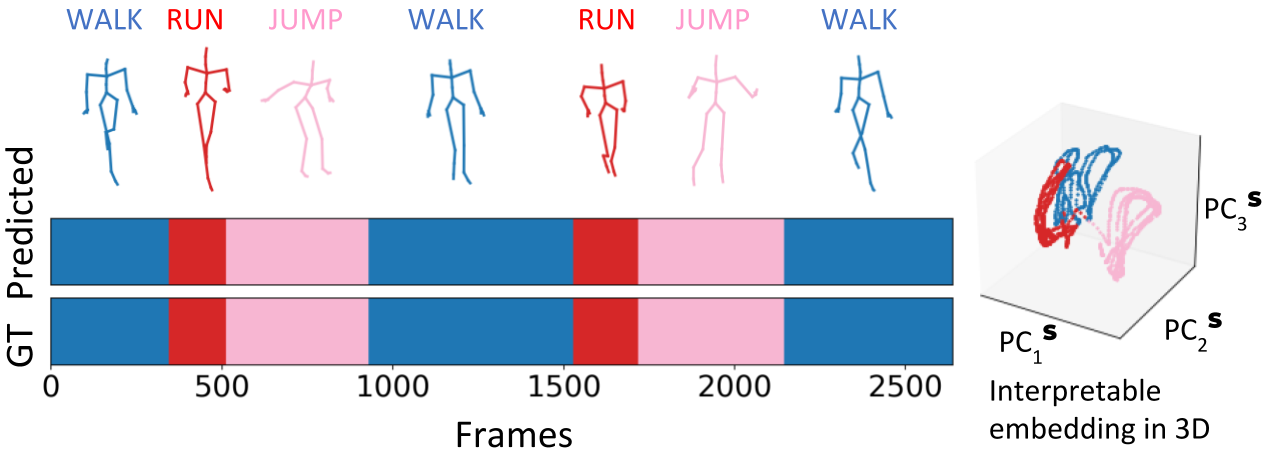}
    \caption{Interpretable embedding of decoder states facilitates \textbf{unsupervised action clustering} of a sequence of human activities with varying durations and individuals.}
    \label{fig:example}
\end{figure}

To demonstrate the generality and a practical application for our methodology, we utilize the embedding space of RNN Seq2Seq and its clustering property to perform \textbf{unsupervised action recognition} on CMU human body motion capture dataset. We evaluate our methods in two cases: (i) we randomly choose sequences from up to eight different actions (walking, running, jumping, soccer, basketball, wash window, directing traffic and basketball signal) and concatenate them together (up to $14400$ frames, $2$ mins), following the same rule from \cite{Li_2018_CVPR}, see Fig.\ref{fig:example}. (ii) Since manually concatenated sequences have discontinuity in the prediction, in the second case we choose trials $1$ to $14$ from subject $86$\footnote{http://mocap.cs.cmu.edu/search.php?subjectnumber=86}, where each sequence contains continuous multiple actions with variable duration. On average each sequence contains 8000 frames and each action sequence has human manually annotated time segments \cite{barbivc2004segmenting}. For each of the two cases, we train various RNN Seq2Seq predictive models to predict a future sequence task with no usage of labels in training. Our results indicate that the Adversarial Geometry-Aware Encoder-Decoder (AGED) model \cite{gui2018adversarial}, shown to be one of the state-of-art motion prediction model based on RNN Seq2Seq as predictor, obtains the best prediction results. We therefore apply our embedding approach in conjunction with the trained AGED model to perform action recognition. Specifically, all sequences which include variety of movements, composed using approach (i) or (ii), are scanned with AGED forward propagation performed to collect the decoder states. The decoder states are then projected to the embedding space. Agglomerative clustering with single linkage and cosine similarity is then used to generate clustering (labeling similar segments to belong to a cluster) . With the the embedding space and decoder attractors within it, \textbf{unsupervised activity recognition} of sequences can be performed. The only information required is the number of unique actions (clusters) that we would like to obtain (see Fig.~\ref{fig:example} and supplementary materials for examples). Testing the algorithm on composed produces the following performance: in case (i), the approach achieves above $98\%$ frame-level accuracy on average; in case (ii), we evaluate our accuracy on by frame-level as well and compare with previously reported methods~\cite{Clopton2017TemporalSC}. Our method reaches an accuracy of $94.75\%$  on average and outperforms previous methods \ref{tab:recognition}.

\begin{table}[]
\label{tab:recognition}
\small
\centering
\caption{Clustering Comparison}
\begin{tabular}{|c|c|}
\hline
\textbf{Method}        & \textbf{Accuracy (\%)} \\ \hline
LRR \cite{liu2010robust}           & 65.08    \\ \hline
SSC \cite{elhamifar2009sparse}           & 76.65    \\ \hline
ACA \cite{zhou2008aligned}           & 84.50    \\ \hline 
TSSC-LC \cite{Clopton2017TemporalSC}       & 86.08    \\\hline \hline
\textbf{Ours} & \textbf{94.75} \\
\hline
\end{tabular}
\end{table}

\subsection{Conclusion}
We propose a novel construction of interpretable embedding for the hidden states of the Seq2Seq model. The embedding clarifies the role of the encoder and the decoder components of Seq2Seq such that encoder embedded trajectories direct the evolution from the origin to the decoder trajectories represented as attractors. Our findings indicate a remarkable property of  networks: the network trained to predict the future evolution of a sequence self-organizes the hidden units representation into separate identities. The identities are revealed through our proposed embedding and clustering. We demonstrate the construction and the utilization of the embedding space on both synthetic and human body joints datasets. We show that the embedding can inspect training and determine the goodness of fit. Furthermore, we present an algorithm for unsupervised clustering of any spatio-temporal features. It utilizes training of Seq2Seq to predict future actions and analyzes the learned representation with the interpretable embedding to generate clustering. We show that such approach allows to perform action recognition on human human body pose data, i.e. to generate an unsupervised time-segmentation and clustering of human movements. The algorithm achieves significantly more robust performance and accuracy than previously proposed approaches. 
\bibliography{ref}

\begin{thebibliography}{}

\bibitem[\protect\citeauthoryear{Alain and
  Bengio}{2016}]{alain2016understanding}
Alain, G., and Bengio, Y.
\newblock 2016.
\newblock Understanding intermediate layers using linear classifier probes.
\newblock {\em arXiv preprint arXiv:1610.01644}.

\bibitem[\protect\citeauthoryear{Arthur and Vassilvitskii}{2007}]{arthur2007k}
Arthur, D., and Vassilvitskii, S.
\newblock 2007.
\newblock k-means++: The advantages of careful seeding.
\newblock In {\em Proceedings of the eighteenth annual ACM-SIAM symposium on
  Discrete algorithms},  1027--1035.
\newblock Society for Industrial and Applied Mathematics.

\bibitem[\protect\citeauthoryear{Barbi{\v{c}} \bgroup et al\mbox.\egroup
  }{2004}]{barbivc2004segmenting}
Barbi{\v{c}}, J.; Safonova, A.; Pan, J.-Y.; Faloutsos, C.; Hodgins, J.~K.; and
  Pollard, N.~S.
\newblock 2004.
\newblock Segmenting motion capture data into distinct behaviors.
\newblock In {\em Proceedings of Graphics Interface 2004},  185--194.
\newblock Canadian Human-Computer Communications Society.

\bibitem[\protect\citeauthoryear{Cho \bgroup et al\mbox.\egroup
  }{2014}]{cho2014learning}
Cho, K.; Van~Merri{\"e}nboer, B.; Gulcehre, C.; Bahdanau, D.; Bougares, F.;
  Schwenk, H.; and Bengio, Y.
\newblock 2014.
\newblock Learning phrase representations using rnn encoder-decoder for
  statistical machine translation.
\newblock {\em arXiv preprint arXiv:1406.1078}.

\bibitem[\protect\citeauthoryear{Clopton \bgroup et al\mbox.\egroup
  }{2017}]{Clopton2017TemporalSC}
Clopton, L.; Mavroudi, E.; Tsakiris, D.~M.; Ali, D.~H.; and Vidal, D.~R.
\newblock 2017.
\newblock Temporal subspace clustering for unsupervised action segmentation.

\bibitem[\protect\citeauthoryear{Collins, Sohl-Dickstein, and
  Sussillo}{2016}]{collins2016capacity}
Collins, J.; Sohl-Dickstein, J.; and Sussillo, D.
\newblock 2016.
\newblock Capacity and trainability in recurrent neural networks.
\newblock {\em arXiv preprint arXiv:1611.09913}.

\bibitem[\protect\citeauthoryear{Elhamifar and
  Vidal}{2009}]{elhamifar2009sparse}
Elhamifar, E., and Vidal, R.
\newblock 2009.
\newblock Sparse subspace clustering.
\newblock In {\em 2009 IEEE Conference on Computer Vision and Pattern
  Recognition},  2790--2797.
\newblock IEEE.

\bibitem[\protect\citeauthoryear{Farrell \bgroup et al\mbox.\egroup
  }{2019}]{farrell2019dynamic}
Farrell, M.~S.; Recanatesi, S.; Lajoie, G.; and Shea-Brown, E.
\newblock 2019.
\newblock Dynamic compression and expansion in a classifying recurrent network.
\newblock {\em bioRxiv}  564476.

\bibitem[\protect\citeauthoryear{Foerster \bgroup et al\mbox.\egroup
  }{2017}]{foerster2017input}
Foerster, J.~N.; Gilmer, J.; Sohl-Dickstein, J.; Chorowski, J.; and Sussillo,
  D.
\newblock 2017.
\newblock Input switched affine networks: An rnn architecture designed for
  interpretability.
\newblock In {\em Proceedings of the 34th International Conference on Machine
  Learning-Volume 70},  1136--1145.
\newblock JMLR. org.

\bibitem[\protect\citeauthoryear{Fragkiadaki \bgroup et al\mbox.\egroup
  }{2015}]{fragkiadaki2015recurrent}
Fragkiadaki, K.; Levine, S.; Felsen, P.; and Malik, J.
\newblock 2015.
\newblock Recurrent network models for human dynamics.
\newblock In {\em Proceedings of the IEEE International Conference on Computer
  Vision},  4346--4354.

\bibitem[\protect\citeauthoryear{Graves, Mohamed, and
  Hinton}{2013}]{graves2013speech}
Graves, A.; Mohamed, A.-r.; and Hinton, G.
\newblock 2013.
\newblock Speech recognition with deep recurrent neural networks.
\newblock In {\em 2013 IEEE international conference on acoustics, speech and
  signal processing},  6645--6649.
\newblock IEEE.

\bibitem[\protect\citeauthoryear{Gui \bgroup et al\mbox.\egroup
  }{2018}]{gui2018adversarial}
Gui, L.-Y.; Wang, Y.-X.; Liang, X.; and Moura, J.~M.
\newblock 2018.
\newblock Adversarial geometry-aware human motion prediction.
\newblock In {\em Proceedings of the European Conference on Computer Vision
  (ECCV)},  786--803.

\bibitem[\protect\citeauthoryear{Hochreiter and
  Schmidhuber}{1997}]{hochreiter1997long}
Hochreiter, S., and Schmidhuber, J.
\newblock 1997.
\newblock Long short-term memory.
\newblock {\em Neural computation} 9(8):1735--1780.

\bibitem[\protect\citeauthoryear{Ionescu \bgroup et al\mbox.\egroup
  }{2014}]{ionescu2014human3}
Ionescu, C.; Papava, D.; Olaru, V.; and Sminchisescu, C.
\newblock 2014.
\newblock Human3. 6m: Large scale datasets and predictive methods for 3d human
  sensing in natural environments.
\newblock {\em IEEE transactions on pattern analysis and machine intelligence}
  36(7):1325--1339.

\bibitem[\protect\citeauthoryear{Jain \bgroup et al\mbox.\egroup
  }{2016}]{jain2016structural}
Jain, A.; Zamir, A.~R.; Savarese, S.; and Saxena, A.
\newblock 2016.
\newblock Structural-rnn: Deep learning on spatio-temporal graphs.
\newblock In {\em Proceedings of the IEEE Conference on Computer Vision and
  Pattern Recognition},  5308--5317.

\bibitem[\protect\citeauthoryear{Karpathy, Johnson, and
  Fei-Fei}{2015}]{karpathy2015visualizing}
Karpathy, A.; Johnson, J.; and Fei-Fei, L.
\newblock 2015.
\newblock Visualizing and understanding recurrent networks.
\newblock {\em arXiv preprint arXiv:1506.02078}.

\bibitem[\protect\citeauthoryear{Kim \bgroup et al\mbox.\egroup
  }{2017}]{kim2017interpretability}
Kim, B.; Wattenberg, M.; Gilmer, J.; Cai, C.; Wexler, J.; Viegas, F.; and
  Sayres, R.
\newblock 2017.
\newblock Interpretability beyond feature attribution: Quantitative testing
  with concept activation vectors (tcav).
\newblock {\em arXiv preprint arXiv:1711.11279}.

\bibitem[\protect\citeauthoryear{Li \bgroup et al\mbox.\egroup
  }{2018}]{Li_2018_CVPR}
Li, C.; Zhang, Z.; Sun~Lee, W.; and Hee~Lee, G.
\newblock 2018.
\newblock Convolutional sequence to sequence model for human dynamics.
\newblock In {\em The IEEE Conference on Computer Vision and Pattern
  Recognition (CVPR)}.

\bibitem[\protect\citeauthoryear{Liu, Lin, and Yu}{2010}]{liu2010robust}
Liu, G.; Lin, Z.; and Yu, Y.
\newblock 2010.
\newblock Robust subspace segmentation by low-rank representation.
\newblock In {\em ICML}, volume~1, ~8.

\bibitem[\protect\citeauthoryear{Luong, Pham, and
  Manning}{2015}]{luong2015effective}
Luong, M.-T.; Pham, H.; and Manning, C.~D.
\newblock 2015.
\newblock Effective approaches to attention-based neural machine translation.
\newblock {\em arXiv preprint arXiv:1508.04025}.

\bibitem[\protect\citeauthoryear{Maaten and
  Hinton}{2008}]{maaten2008visualizing}
Maaten, L. v.~d., and Hinton, G.
\newblock 2008.
\newblock Visualizing data using t-sne.
\newblock {\em Journal of machine learning research} 9(Nov):2579--2605.

\bibitem[\protect\citeauthoryear{Martinez, Black, and
  Romero}{2017}]{martinez2017human}
Martinez, J.; Black, M.~J.; and Romero, J.
\newblock 2017.
\newblock On human motion prediction using recurrent neural networks.
\newblock In {\em Proceedings of the IEEE Conference on Computer Vision and
  Pattern Recognition},  2891--2900.

\bibitem[\protect\citeauthoryear{Recanatesi \bgroup et al\mbox.\egroup
  }{2018}]{recanatesi2018signatures}
Recanatesi, S.; Farrell, M.; Lajoie, G.; Deneve, S.; Rigotti, M.; and
  Shea-Brown, E.
\newblock 2018.
\newblock Signatures and mechanisms of low-dimensional neural predictive
  manifolds.
\newblock {\em bioRxiv}  471987.

\bibitem[\protect\citeauthoryear{Shlizerman, Schroder, and
  Kutz}{2012}]{shlizerman2012neural}
Shlizerman, E.; Schroder, K.; and Kutz, J.~N.
\newblock 2012.
\newblock Neural activity measures and their dynamics.
\newblock {\em SIAM Journal on Applied Mathematics} 72(4):1260--1291.

\bibitem[\protect\citeauthoryear{Strobelt \bgroup et al\mbox.\egroup
  }{2019}]{strobelt2019s}
Strobelt, H.; Gehrmann, S.; Behrisch, M.; Perer, A.; Pfister, H.; and Rush,
  A.~M.
\newblock 2019.
\newblock S eq 2s eq-v is: A visual debugging tool for sequence-to-sequence
  models.
\newblock {\em IEEE transactions on visualization and computer graphics}
  25(1):353--363.

\bibitem[\protect\citeauthoryear{Sutskever, Vinyals, and
  Le}{2014}]{sutskever2014sequence}
Sutskever, I.; Vinyals, O.; and Le, Q.~V.
\newblock 2014.
\newblock Sequence to sequence learning with neural networks.
\newblock In {\em Advances in neural information processing systems},
  3104--3112.

\bibitem[\protect\citeauthoryear{Zeiler and
  Fergus}{2014}]{zeiler2014visualizing}
Zeiler, M.~D., and Fergus, R.
\newblock 2014.
\newblock Visualizing and understanding convolutional networks.
\newblock In {\em European conference on computer vision},  818--833.
\newblock Springer.

\bibitem[\protect\citeauthoryear{Zhou, De~la Torre, and
  Hodgins}{2008}]{zhou2008aligned}
Zhou, F.; De~la Torre, F.; and Hodgins, J.~K.
\newblock 2008.
\newblock Aligned cluster analysis for temporal segmentation of human motion.
\newblock In {\em 2008 8th IEEE international conference on automatic face \&
  gesture recognition},  1--7.
\newblock IEEE.

\end{thebibliography}
\bibliographystyle{aaai.bst}

\end{document}